\documentclass[runningheads]{llncs}

\usepackage[english]{babel}
\usepackage[utf8x]{inputenc}
\usepackage[T1]{fontenc}

\usepackage{graphicx}
\usepackage{amssymb}
\usepackage{amsmath}
\usepackage[colorinlistoftodos]{todonotes}
\usepackage[colorlinks=true, allcolors=blue]{hyperref}
\usepackage{stackengine}
\usepackage{dsfont}

\DeclareMathOperator*{\probmapp}{\mathcal{M}}

\title{A unified view on differential privacy and robustness to adversarial examples}

\author{Rafael Pinot\inst{1,2} \and
Florian Yger\inst{1} \and
C\'edric Gouy-Pailler\inst{2} \and Jamal Atif \inst{1}}
\authorrunning{R. Pinot et al.}

\institute{Universit\'e Paris-Dauphine, PSL Research University, LAMSADE, Paris, France \and
Institut LIST, CEA, Universit\'e Paris-Saclay, LADIS, Palaiseau, France
\email{rafael.pinot@dauphine.fr}\\}

\begin{document}
\maketitle

\begin{abstract}
This short note highlights some links between two lines of research within the emerging topic of trustworthy machine learning: differential privacy and robustness to adversarial examples. By abstracting the definitions of both notions, we show that they build upon the same theoretical ground and hence results obtained so far in one domain can be transferred to the other. More precisely, our analysis is based on two key elements: probabilistic mappings (also called randomized algorithms in the differential privacy community), and the Renyi divergence which subsumes a large family of divergences. We first generalize the definition of robustness against adversarial examples to encompass probabilistic mappings. Then we observe that Renyi-differential privacy (a generalization of differential privacy recently proposed in~\cite{Mironov2017RenyiDP}) and our definition of robustness share several similarities. We finally discuss how can both communities benefit from this connection to transfer technical tools from one research field to the other.  
\end{abstract}

\keywords{Differential Privacy \and Adversarial Examples \and Renyi divergence}

\section{Introduction}

With the large adoption of machine learning techniques in several domains (including critical ones), researchers and practitioners are observing growing concerns on the security and privacy of the tools they develop. A primary concern is to guarantee that sensitive information from the used databases are not leaked, accidentally disclosed, or inferred from the sole release of the model (\emph{privacy preserving algorithms}). Beyond preserving privacy, a crucial issue of recent machine learning approaches is to protect the methods against malicious users targeting their weaknesses (e.g \emph{adversarial examples}, or poisoning attacks). 
\paragraph{Privacy preserving algorithms:} Several definitions have been introduced to characterize privacy preserving algorithms in the context of machine learning and data publishing. Among them, differential privacy has become the dominant standard by providing a formal and adaptive conception of privacy preserving data-analysis. It has been broadly investigated in numerous frameworks and applications (see \cite{Dwork_2013} for a complete overview of the field). An algorithm is said to be differentially private if, given two close databases, it produces statistically indistinguishable outputs. Highly correlated to the notion of "closeness" both in the input and output spaces, most frameworks~\cite{Dwork_2013,Chatzikokolakis2013,ElSalamouny2014,DworkR16} rely on divergences/pseudo-metrics between probability measures to characterize this notion. Recently, Mironov~\cite{Mironov2017RenyiDP} proposed to use the well-known Renyi divergence to obtain a more general definition of privacy. This notion is well defined, and it  exhibits principled theoretical advantages over previous definitions, which makes it the most general formulation of differential privacy introduced so far. 

\paragraph{Adversarial examples attacks:} Modern neural networks achieve state of the art performances in a variety of domains. However, it has been shown that such neural networks can be vulnerable to adversarial examples, i.e.  imperceptible variations of legitimate examples crafted to deliberately mislead a machine learning algorithm~\cite{Szegedy2013IntriguingPO}. Since then, attacks and defenses are developed in a tight back-and-forth(see~\cite{OnEvaluatingrobustness} for a complete overview of the field). Most past defenses were deterministic (see e.g~\cite{Madry2018TowardsDL,Samangouei2018DefenseGAN}), but recently, the idea of using randomization in the learning process to ensure robustness against adversarial examples attacks is gaining in interest~\cite{lecuyer2018certified,KolterRandomizedsmoothing,PinotRobustnessExponential}.

\paragraph{Outline of the paper: } We first recall the key notions of probabilistic mapping and Renyi divergence in Section~\ref{section:preliminaries}. Then we introduce the notion of differential privacy and present its generalization called Renyi-differential privacy in Section~\ref{section::DP}. Section~\ref{section:robustness} presents the problem of adversarial examples and our generalized definition of robustness to these attacks. Finally we discuss in Section~\ref{section::linkbetweenDPandrobustness} the similarity between the two concepts, and an application to image classification in which we transfer tools from differential privacy to make algorithms robust to adversarial examples.

\section{Preliminaries}
\label{section:preliminaries}

Let us consider two arbitrary metric spaces $(\mathcal{X},d_{\mathcal{X}})$, and $(\mathcal{Y},d_{\mathcal{Y}})$, let $\sigma(\mathcal{Y})$ be a $\sigma \small{-} algebra$ over $\mathcal{Y}$ and $\mathcal{P}(\mathcal{Y})$ be the set of probability measures over $(\mathcal{Y},\sigma(\mathcal{Y}))$. The notion of probabilistic mapping is the central concept used in differential privacy, we recall it below.

\begin{definition}[probabilistic mapping]
A probabilistic mapping from $\mathcal{X}$ to $\mathcal{Y}$ is a mapping $\probmapp: \mathcal{X} \to \mathcal{P}(\mathcal{Y})$. Given $x$, $\probmapp$ outputs a probability measure $\probmapp(x)$. To get a numerical output $y$ out of $\probmapp$ for $x$, one needs to sample $y\sim \probmapp(x)$.
\end{definition}

\noindent Informally, a probabilistic mapping $\probmapp$ is said to be differentially private, if given $x$ and $x'$ two close inputs (i.e $d_{\mathcal{X}}(x,x')$ is small enough) it outputs two close measures $\probmapp(x)$, and $\probmapp(x')$. To evaluate the closeness between this two probability measures in the formal definition of differential privacy, Dwork et. al~\cite{Dwork_2013} uses the maximum divergence, which is a special case of the more general Renyi divergence defined as follows: 

\begin{definition}[Renyi divergence of order $\lambda$~\cite{renyi1961}]
Let us consider $\mu_1$, $\mu_2$ $\in$ $\mathcal{P}(\mathcal{Y})$ two probability measures, both dominated by a third measure $\nu$. The Renyi divergence of order $\lambda$ between $\mu_1$ and $\mu_2$ writes 
$$D_{\lambda}(\mu_1,\mu_2):=\frac{1}{\lambda -1}\log \int_{\mathcal{Y}} g_2(y)  \left(\frac{g_1(y)}{g_2(y)}\right)^{\lambda} d\nu(y). $$
Where $g_1$ and $g_2$ are the probability density of $\mu_1$, and $\mu_2$ with respect to $\nu$. 
\end{definition}

\noindent The Renyi divergence (see~\cite{Renyi-Kullback-Link} for more details) is defined for $\lambda \in (1,\infty)$. It equals the Kullback-Leibler divergence when $\lambda \rightarrow 1$, and the maximum divergence (denoted $D_{\infty}$) when $\lambda \rightarrow \infty$.  
It also has the very special property of being non decreasing with respect to $\lambda$. This divergence is very common in machine learning (especially the Kullback-Leibler divergence), statistics, and information theory. Using this notion of closeness between distributions, one can define both differential privacy (with $D_{\infty}$), and Renyi-differential privacy (with $D_{\lambda}$).

\section{Differential privacy and its generalization}
\label{section::DP}
We now present the definition of differential privacy, and its Renyi generalization.

\begin{definition}[Classical differential privacy~\cite{Dwork_2013}]
Let $\mathcal{X}$ be a space of databases, $\mathcal{Y}$ an output space, and "$\sim_h$" denoting the that two databases from  $\mathcal{X}$ only differ from one row. A probabilistic mapping $\probmapp$ from $\mathcal{X}$ to $\mathcal{Y}$ is called differentially private if for any $x, x' \in \mathcal{X} \textnormal{ s.t. } x \sim_h x'$ $\textnormal{ and for any } Y \in \sigma(\mathcal{Y})$ $\textnormal{ on has } \probmapp(x)(Y) \leq \exp(\epsilon)\probmapp(x')(Y).$
\end{definition}

\begin{definition}[Metric differential privacy~\cite{Chatzikokolakis2013}]\label{definition:generalizeDP}
Let $\epsilon>0$, $(\mathcal{X},d_{\mathcal{X}})$ an arbitrary (input) metric space, and $\mathcal{Y}$ an output space. A probabilistic mapping $\probmapp$ from $\mathcal{X}$ to $\mathcal{Y}$ is called $(\epsilon,\alpha)$-$d_{\mathcal{X}}$ private if for any $ x, x'  \textnormal{ s.t } d_{\mathcal{X}}(x,x') \leq \alpha$, $\textnormal{ one has } D_{\infty}\left(\probmapp(x),\probmapp(x')\right) \leq \epsilon. $
\end{definition}

\noindent \emph{Classical differential privacy} is a particular case of \emph{Metric differential privacy} where $\mathcal{X}$ is a set of tabular databases, $d_{\mathcal{X}}$ is the hamming distance, and $\alpha =1$\footnote{Classical definitions set $\alpha=1$, and argue that one can always scale $d_{\mathcal{X}}$ such that $d_{\mathcal{X}} \leq 1$ fits the notion of "close enough". We rather keep $d_{\mathcal{X}}$ unchanged and take an arbitrary $\alpha$ instead. Both definitions are equivalent.}. We finally introduce a general form of privacy definition that complies both with classical, and metric differential privacy, namely Renyi-differential privacy.

\begin{definition}[Renyi differential privacy~\cite{Mironov2017RenyiDP}]\label{definition:renyiDP}
Let $\epsilon>0$, $(\mathcal{X},d_{\mathcal{X}})$ an arbitrary (input) metric space, and $\mathcal{Y}$ the output space. A probabilistic mapping $\probmapp$ from $\mathcal{X}$ to $\mathcal{Y}$ is called $(\lambda,\epsilon,\alpha)$-$d_{\mathcal{X}}$ Renyi-private if for any $ x, x'  \textnormal{ s.t } d_{\mathcal{X}}(x,x') \leq \alpha$, $\textnormal{ one has } D_{\lambda}\left(\probmapp(x),\probmapp(x')\right) \leq \epsilon $
\end{definition}

\noindent According to Definition~\ref{definition:renyiDP}, it is clear that both Metric, and differential privacy are included in Renyi-differential privacy.
Moreover, note that the definition above is based on arbitrary spaces, and metrics ($\mathcal{X}$, $d_{\mathcal{X}}$, and $\mathcal{Y}$). Hence, one can define Renyi-privacy for an arbitrary learning task, even if preserving privacy in this task has no clear semantic. In the following, we present robustness against adversarial examples, and how robustness and privacy are formally similar.


\section{Robustness to adversarial examples}
\label{section:robustness}

Let us now consider a classification task over $\mathcal{X}$ (i.e $\mathcal{Y}=[N]$). Let us denote $\mathcal{D}$ the ground-truth distribution one tries to learn, and $h$ the classifier at hand  (trained over some subset of $\mathcal{X}\times \mathcal{Y}$). An adversarial example attack for $x$ is a small perturbation of $x$ that fools the results of $h$. For instance, for image classification, the changes from the initial image to the perturbed one are visually imperceptible, but images are classified with two different labels. The problem of generating an adversarial example from an input $x$ writes
\begin{align}
\min d_{\mathcal{X}}\left(x,x+\tau \right), \textnormal{ where $\tau \in \mathcal{X} ,\textnormal{ and } h(x+\tau)\neq h(x)$}\label{attackclassicalproblem}
\end{align}
\noindent Even if adversarial examples are intensively studied, a broadly accepted definition of robustness against adversarial attacks does not seem to exist. We settle that the notion of prediction-change risk initially formalized in~\cite{NIPS2018Mahloujifar}, and implicitly used in e.g~\cite{Szegedy2013IntriguingPO} is a suitable start-point. Given a classifier $h$, it is defined as
$$\mathbb{P}_{x \sim \mathcal{D}_{\mathcal{X}}}\left[ \exists x' \in B(x,\alpha) \text{ s.t } h(x')\neq h(x) \right].$$ Where $B(x,\alpha) =\{x' \in \mathcal{X} \text{ s.t } d_{\mathcal{X}}(x,x') \leq \alpha \}$, and $\mathcal{D}_{\mathcal{X}}$ is the marginal distribution of $\mathcal{D}$ with respect to $\mathcal{X}$. From this we can derive a definition of robustness to adversarial attacks.

\begin{definition}[Adversarial robustness]
\label{def::ClassicalRobustness} A classifier $h$ is said to be $(\alpha,\gamma)$-robust if $\mathbb{P}_{x \sim \mathcal{D}_{\mathcal{X}}}\left[ \exists x' \in B(x,\alpha) \text{ s.t } h(x')\neq h(x) \right] \leq \gamma.$
\end{definition}

 \noindent Regarding~\cite{Xie2017MitigatingAE,MoosaviDezfooli2018DivideDA}, probabilistic mappings seem to be good candidates to defend against adversarial example attacks. The following definition gives a generalized notion of robustness against adversarial examples attacks complying with probabilistic mappings. 

\begin{definition}[Generalized adversarial robustness]
\label{definition:generalizedrobustness}
Let $D_{\mathcal{P}(\mathcal{Y})}$ be a metric/divergence on $\mathcal{P}(\mathcal{Y})$. A randomized classifier $\probmapp$ is said to be $D_{\mathcal{P}(\mathcal{Y})}$-$(\alpha, \epsilon, \gamma)$-robust if
$\mathbb{P}_{x\sim \mathcal{D}_{\mathcal{X}}}\left[ \exists x' \in B(x,\alpha) \text{ s.t } D_{\mathcal{P}(\mathcal{Y})}(\probmapp(x'),\probmapp(x)) > \epsilon \right] \leq \gamma. $
\end{definition}

\noindent Definition~\ref{definition:generalizedrobustness} is fully general, and depends on the metric/divergence $D_{\mathcal{P}(\mathcal{Y})}$ one chooses to consider. In particular, if one restricts the study of randomized classifiers to Dirac measures, and sets $D_{\mathcal{P}(\mathcal{Y})}$ to be the trivial distance (which takes the value $0$ where the measures are equal and $1$ elsewhere), definitions~\ref{def::ClassicalRobustness} and~\ref{definition:generalizedrobustness} match. One can refer to~\cite{PinotRobustnessExponential} for more details on definition~\ref{definition:generalizedrobustness} and proof on the interest of choosing $D_{\mathcal{P}(\mathcal{Y})}$ to be the Renyi divergence $D_{\lambda}$.

\section{Links between differential privacy and robustness to adversarial attacks}
\label{section::linkbetweenDPandrobustness}

The starting point to highlight the similarities between both notions are Definitions~\ref{definition:renyiDP} and~\ref{definition:generalizedrobustness}. A first observation is that in an abstract way (i.e. without instantiating the spaces), and by considering the Renyi divergence both definitions are strictly equivalent. This suggests the following claim, the proof of which is straightforward since it follows from the definitions.


\begin{claim}[Renyi-DP $\iff$ $D_{\lambda}$-robustness]\label{mainclaim}
An algorithm $\probmapp$ is $D_{\lambda}$-$(\alpha, \epsilon, 0)$-robust if and only if $\probmapp$ is $\mathcal{D}_{\mathcal{X}}$-almost surely $(\lambda,\epsilon,\alpha)$-$d_{\mathcal{X}}$ Renyi-differentially private.
\end{claim}

\noindent While this mathematical equivalence is important from a theoretical point of view, we will now go into deeper details to consider practical implications of this formulation. Without loss of generality, practical settings, in which privacy or robustness are needed, can be classified into three categories:

\paragraph{1.} Differential privacy and adversarial robustness need to be ensured: prominent examples of this situation are image or voice classification. In this case, instead of considering two separate methodologies, both problems can be treated simultaneously, with the same tools.

\paragraph{2.} Adversarial robustness has to be ensured but there are no special constraint on privacy: in this case, thanks to Claim~\ref{mainclaim}, one could be able to design new defense mechanisms against adversarial examples attacks based on the extensive literature on differential privacy. Accordingly, one can design new defense mechanisms against adversarial example attacks based on the noise injection techniques traditionally used in the differential privacy literature proposed in e.g~\cite{PinotRobustnessExponential,lecuyer2018certified,KolterRandomizedsmoothing}. Note that, even though the formal connecting between differential privacy and robustness to adversarial examples is not identically stated in~\cite{lecuyer2018certified,KolterRandomizedsmoothing} and this note, both visions are not conflicting.

\paragraph{3.} Differential privacy needs to be ensured but robustness to adversarial examples is not needed: while this setting does not seem intuitively natural, we advocate that a few emerging frameworks currently actively developed to test against adversarial robustness could also be used to evaluate differential privacy with minor adaptations. \\

\noindent Point~\textbf{2} is currently being investigated. We however argue that the explicit link we just draw  between differential privacy and robustness to adversarial examples might lead practitioners from both side to investigate further points \textbf{1} and \textbf{3}.

\bibliographystyle{splncs03}
\bibliography{bibliography}

\end{document}